%% file: acl2023.tex
\pdfoutput=1

\documentclass[11pt]{article}

\usepackage[]{ACL2023}

\usepackage{times}
\usepackage{latexsym}

\usepackage[T1]{fontenc}

\usepackage[utf8]{inputenc}

\usepackage{microtype}

\usepackage{inconsolata}

\usepackage{graphicx}
\usepackage{caption}
\usepackage{booktabs}
\usepackage{booktabs}
\usepackage{multirow}
\usepackage{array}
\usepackage[inline]{enumitem}

\newcommand*\rotbf[1]{\rotatebox{45}{#1}}

\title{Effortless Integration of Memory Management into\\ Open-Domain Conversation Systems}

\author{Eunbi Choi$^{1}$\thanks{*Work done during internship at Kakao Brain.} \; Kyoung-Woon On$^2$ \; {\bf Gunsoo Han$^2$} \; {\bf Sungwoong Kim$^{3}$} \\ {\bf Daniel Wontae Nam$^2$} \; {\bf Daejin Jo$^2$} \; {\bf Seung Eun Rho$^2$} \; {\bf Taehwan Kwon$^2$} \; {\bf Minjoon Seo$^1$} \\
  $^1$KAIST AI \;\;\;\; $^2$Kakao Brain \;\;\;\; $^3$Korea University \\
  \texttt{\{eunbi,minjoon\}@kaist.ac.kr} \\
  \texttt{\{kloud.ohn,coco.han,dwtnam,daejin.jo\}@kakaobrain.com} \\
  \texttt{swkim01@korea.ac.kr} \; \texttt{seungeun07@snu.ac.kr} \; \texttt{kwonth0315@gmail.com} \\
  }

\begin{document}
\maketitle

\input{Sections/0_abstract.tex}
\input{Sections/1_introduction.tex}

\input{Sections/2_related_work.tex}
\input{Sections/3_method.tex}
\input{Sections/4_experiments.tex}

\input{Sections/5_conclusion.tex}
\input{Sections/7_ethic.tex}

\bibliography{anthology,custom}
\bibliographystyle{acl_natbib}

\appendix



\end{document}

%% file: Sections/0_abstract.tex
\begin{abstract}



Open-domain conversation systems integrate multiple conversation skills into a single system through a modular approach. 
One of the limitations of the system, however, is the absence of management capability for external memory.
In this paper, we propose a simple method to improve \textsc{BlenderBot3} by integrating memory management ability into it.
Since no training data exists for this purpose, we propose an automating dataset creation for memory management.
Our method 1) requires little cost for data construction, 2) does not affect performance in other tasks, and 3) reduces external memory.
We show that our proposed model \textsc{BlenderBot3-M\^{}3}
, which is multi-task trained with memory management, outperforms \textsc{BlenderBot3} with a relative 4\% performance gain in terms of F1 score. 
\end{abstract}

%% file: Sections/1_introduction.tex
\section{Introduction}
\label{sec:intro}
Open-domain conversation, which refers to conversing without any constraints on topic (e.g., chitchat), has been the subject of active research in recent years~\citep{shuster2022blenderbot,Roller2020RecipesFB,Thoppilan2022LaMDALM,DeFreitas2020TowardsAH}.
A good open-domain conversational agent is expected to be engaging, knowledgeable, up-to-date, and personalized by remembering the user. 
Therefore it is key to seamlessly blend all desirable skills into a conversation system.

To address this, previous works utilized separate modules for internet search~\citep{shuster2022language,komeili-etal-2022-internet} or memory generation~\citep{zhong-etal-2022-less,xu-etal-2022-long,xu-etal-2022-beyond}.
Recently, there have been studies aimed at unifying various conversation abilities into a single language model based on the modular approach~\citep{shuster2022language,shuster2022blenderbot}.
One notable recent work is \textsc{BlenderBot3}~\citep{shuster2022blenderbot}, a modular system where a single transformer model is served for all modules.

Recently, there is a trend of providing personalized conversation experiences by memorizing individual user information~\citep{xu-etal-2022-long,Lu2022PartnerPG,Bae2022KeepMU,Mazar2018TrainingMO}. 
However, as shown in Figure~\ref{fig:1}, the agent may encounter problems when the conversation lasts long, as information about a person is not static and changes over time.
Therefore, managing memory based on the current state is one of key abilities for a good open domain conversational agent~\citep{Bae2022KeepMU}.


In this paper, we propose an effortless method to improve \textsc{BlenderBot3} by integrating memory management capability into it.
Since no general data exists for this purpose, we formally define a new task for memory management and present an automated method to create memory management datasets.
Our method has following advantages:
\begin{itemize}
    \item Require little cost for data construction.
    \item Do not affect \textsc{BlenderBot3}'s performance in other tasks.
    \item Need no additional costs for the external memory and model parameters, but rather reduces the costs.
\end{itemize}
We leverage publicly available datasets to construct memory management data, which can be easily scaled up and extended to other domains.
Additionally, we report performance in other tasks of \textsc{BlenderBot3} and external memory efficiency of our model.

\input{figure/figure.tex}

Experimental results show that our proposed model \textsc{BlenderBot3-M\^{}3}, which is multi-task trained with memory management, outperforms \textsc{BlenderBot3} with a relative 4\% performance gain in terms of F1 score. 
In addition, across all 67 tasks where \textsc{BlenderBot3} is trained, we observe that the average PPL score of \textsc{BlenderBot3-M\^{}3} increases 0.05 from the PPL of \textsc{BlenderBot3}, demonstrating the seamless integration of the new memory management task.
Through these explorations, we demonstrate that the overall conversation performance is effortlessly, yet successfully improved by incorporating the memory management capability into the conversational agent.


%% file: figure/figure.tex
\begin{figure*} 
    \includegraphics[width=1.0\textwidth]{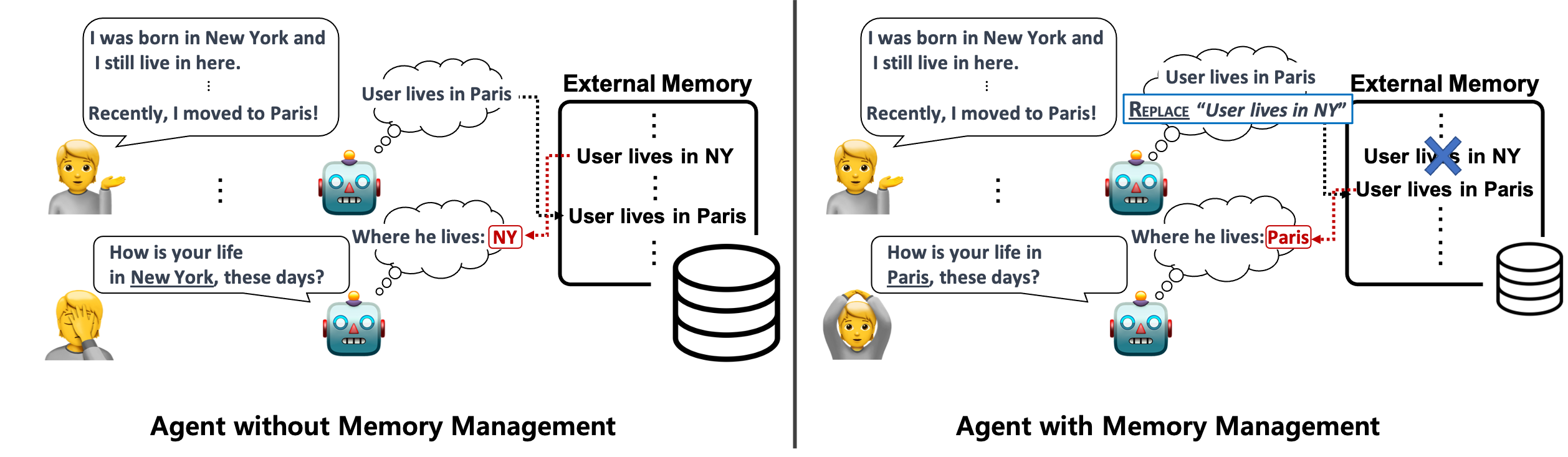}
    \caption{Illustrative examples of open-domain conversation with an agent with/without memory management. \textit{Left}: Without memory management, an agent cannot handle the situation of changing user's information. Also, the size of the memory is monotonically increased during the conversation. \textit{Right}: With memory management, the agent can replaces the out-dated information with the new one. In addition, the size of the memory also tend to be suppressed.
    }
    \label{fig:1}
\end{figure*}

%% file: Sections/2_related_work.tex
\section{Related Work}
\label{sec:related}
\subsection{Unified Conversation Systems}
Recently, there have been studies for unifying different conversation skills, such as being engaging, factual, empathetic, civil and knowledgeable, into a single language model based on modular approach~\citep{Smith2020CanYP,shuster2022language,shuster2022blenderbot,Ung2021SaFeRDialoguesTF,Kim2022ProsocialDialogAP}.
Therefore, it is crucial to incorporate all desirable skills into a conversation system seamlessly.
\textsc{BlenderBot3}~\citep{shuster2022blenderbot} shows that equipping a single transformer model with all skills through multi-task training can be a promising direction.

\subsection{Personalized Conversation Systems with Memory}
Providing personalized conversation experiences to users has been improved by memorizing their information.
Conversational agents either keep the user's profile~\citep{Zhang2018PersonalizingDA} or extracted user information from conversation history~\citep{xu-etal-2022-long,Lu2022PartnerPG,Bae2022KeepMU} to generate personalized response.
Especially, \citet{xu-etal-2022-long} extract and store user information dynamically during conversation, allowing the agents to remember the user in long-term conversations.

\subsection{Memory in Long-term Conversation Systems}
\citet{Xu2021BeyondGM} and \citet{xu-etal-2022-long} have tackled long-term conversation problem and released MSC and DuLeMon datasets, respectively. 
In MSC~\citep{Xu2021BeyondGM} dataset, sessions are annotated with summaries of previous sessions that may be useful in current conversations. 
It is intended to refer to the previous conversation with memory in long-term.
However, MSC does not aim to reflect the dynamic feature of personal information.

Work of \citet{Bae2022KeepMU} represents the first attempt to address this problem, presenting the memory management task.
However, since their approach classifies relationship of sentence pairs to compute memory management operation, the inference time increases as the length of memory increases.

%% file: Sections/3_method.tex
\section{Approach}
\label{sec:approach}

We improve \textsc{BlenderBot3} by equipping it with memory management capability, which requires data of memory management for multi-task training along with other tasks.
Since no general data exists for this purpose, we define a new task for memory management and present an effortless way to create the memory management dataset, which can be easily scaled up. 
In this paper, we apply this method to open-domain conversations, but it is generally extendable to memory management in other tasks.

\subsection{Memory Management Task Definition}
During conversation, a conversational agent maintains natural language memory sentences \(\mathrm{M_t = \{m_1, m_2, \cdots, m_n\}}\) which consists of user information abstracted from the previous utterances.
At time step $\mathrm{t}$, we are given the memory $\mathrm{M_t}$ and user information $\mathrm{p_t}$ generated from utterance $\mathrm{u_t}$, each of which could be either user's utterance or bot's utterance.
At the end of each turn, we aim to predict memory management operation $\mathrm{op}$. We define the management operation set as \(\mathrm{O =\{APPEND, PASS, REPLACE\ m_i\}}\) where $\mathrm{m_i}$ is an entry of the memory $\mathrm{M_t}$ and define this task as \(\mathrm{Model(M_t, p_t) \rightarrow op}\) where \(\mathrm{op \in O}\). 
Consequently, the model is required to determine whether to add $\mathrm{p_t}$, not add $\mathrm{p_t}$, or replace $\mathrm{m_i}$ with $\mathrm{p_t}$ by considering all memory $\mathrm{M_t}$ and $\mathrm{p_t}$ holistically.

\subsection{Memory Management Data Curation}
\label{sec3:creation_method}
For memory management training, we need \(\mathrm{\langle M_t, p_t, op \rangle}\) triples.
As there is no existing dataset providing the triple, we construct it in an automated way with existing datasets.
We reinterpret publicly available DNLI~\citep{Welleck2018DialogueNL} dataset, that is designed for detecting textual entailment in dialogs, for memory management operations.
Given a DNLI triple \(\mathrm{\langle s_1, s_2, relationship \rangle}\), we utilize s1 as a memory sentence to be part of all memory $\mathrm{M_t}$ and $\mathrm{s_2}$ as $\mathrm{p_t}$, newly generated information that is to be added or not.
Then, we reinterpret the relationship labels as memory operations as below:
\begin{itemize}
  \item \textbf{Positive} means $\mathrm{s_1}$ and $\mathrm{s_2}$ share relevant information.
  However, the relationship between $\mathrm{s_1}$ and $\mathrm{s_2}$ can be either $\mathrm{s_1}$ entails $\mathrm{s_2}$, $\mathrm{s_2}$ entails $\mathrm{s_1}$, or almost identical, depending on the amount of information.
  We classify DNLI data with positive labels into the three categories above and label them as $\mathrm{PASS}$, $\mathrm{REPLACE \ s_1}$, and $\mathrm{APPEND}$ operations, respectively.
  \item \textbf{Negative} means $\mathrm{s_1}$ and $\mathrm{s_2}$ are contradictory. We make up $\mathrm{REPLACE \ s_1}$ operations with those data. 
  \item \textbf{Neutral} means $\mathrm{s_1}$ and $\mathrm{s_2}$ are not related. We construct data with $\mathrm{APPEND}$ operation using neutral data.

\end{itemize}

Then, random memory sentences are collected to construct the memory $\mathrm{M'}$ of various lengths.
We append $\mathrm{s_1}$ to $\mathrm{M'}$ to comprise the entire memory $\mathrm{M}$ and finally obtain \(\mathrm{\langle M, s2, op \rangle}\) triples where \(\mathrm{M = M' \cup \{s1\}}\). 


%% file: Sections/4_experiments.tex
\section{Experiments}
\label{sec:exp}

\begin{table*}[t]
    \vspace*{1mm}
    \centering
    \small\addtolength{\tabcolsep}{-1pt}
    \fontsize{10}{11}\selectfont
    \begin{tabular}{lcccccc}
    \toprule
    \multirow{2}{*}{}   & \multicolumn{1}{c}{\textbf{Session 2}}
                        & \multicolumn{1}{c}{\textbf{Session 3}}
                        & \multicolumn{1}{c}{\textbf{Session 4}} 
                        & \multicolumn{1}{c}{\textbf{Session 5}} 
                        & \multicolumn{1}{c}{\textbf{All Sessions}}
                        & \multicolumn{1}{c}{\textbf{Memory}} \\
    \cmidrule(lr){2-2} \cmidrule(lr){3-3} \cmidrule(lr){4-4} \cmidrule(lr){5-5} \cmidrule(lr){6-6} \cmidrule{7-7}
    {} & F1 ($\uparrow$) & F1 ($\uparrow$) & F1 ($\uparrow$) & F1 ($\uparrow$) & F1 ($\uparrow$) & \# of entries ($\downarrow$)\\
    \midrule
    \textsc{BlenderBot3} & \underline{0.1791} & \underline{0.1747} & \underline{0.1528} & \underline{0.149} & \underline{0.164} & 117.7 \\
    \midrule
    \textsc{BlenderBot3} & \multicolumn{5}{c}{} \\
    \quad{+ MM fine-tuning} & 0.1732 & 0.1681 & 0.1443 & 0.1376 & 0.1558 & \underline{85.7}  \\ 
    \midrule
    \textsc{BlenderBot3-M\^{}3 (ours)} & \textbf{0.1882} & \textbf{0.1828} & \textbf{0.1583} & \textbf{0.1536} & \textbf{0.1708} & \textbf{70.3} \\
    \bottomrule
    \end{tabular}
    \caption{\label{tab1:e2e-eval}
    End-to-end evaluation for measuring overall conversation ability on MSC dataset.
    \textsc{BlenderBot3-M\^{}3}, multi-task trained with memory management data, outperforms \textsc{BlenderBot3} and \textsc{BlenderBot3}+ MM fine-tuning across all sessions.
    }
\end{table*}

\begin{table*}[t]
    \centering
    \small\addtolength{\tabcolsep}{-4.5pt}
    \fontsize{8}{11}\selectfont
    \begin{tabular}{l *{13}c}
    \toprule
    {} & \multicolumn{2}{c}{Decision} & \multicolumn{2}{c}{Generation} & \multicolumn{3}{c}{Knowledge} & \multicolumn{4}{c}{Dialogue} & \multicolumn{1}{c}{All} & \multicolumn{1}{c}{Management} \\
    \cmidrule(lr){2-3} \cmidrule(lr){4-5} \cmidrule(lr){6-8} \cmidrule(lr){9-12} \cmidrule(lr){13-13} \cmidrule(lr){14-14}
    {} & \rotbf{Search} & \rotbf{Memory} & \rotbf{Query} & \rotbf{Memory} & \rotbf{Search} & \rotbf{Memory} & \rotbf{Entity} & \rotbf{Search} & \rotbf{Memory} & \rotbf{Entity} & \rotbf{Vanilla} & {} & \rotbf{Memory}\\
    \cmidrule(lr){2-3} \cmidrule(lr){4-5} \cmidrule(lr){6-8} \cmidrule(lr){9-12} \cmidrule(lr){13-13} \cmidrule(lr){14-14}
     {} & \multicolumn{2}{c}{PPL ($\downarrow$)} & \multicolumn{2}{c}{PPL ($\downarrow$)} & \multicolumn{3}{c}{PPL ($\downarrow$)} & \multicolumn{4}{c}{PPL ($\downarrow$)} & \multicolumn{1}{c}{PPL ($\downarrow$)} & \multicolumn{1}{c}{PPL ($\downarrow$)} \\
    \midrule
    \textsc{BlenderBot3} & 1.028 & 1.003 & 5.847 & 2.522 & 1.759 & 1.247 & 5.010 & 2.729  & 9.115 & 10.218 & 11.260 & 3.144 & -\\
    \midrule
    \textsc{BlenderBot3-M\^{}3 (ours)} & 1.028 & 1.003 & 5.627 & 2.567 & 1.798 & 1.257 & 5.405 & 2.781 & 9.181 & 10.371 & 11.485 & 3.196 & 1.735  \\
    \bottomrule
    \end{tabular}
    \caption{\label{tab2:modular-eval}
        Modular performance (PPL) for other conversation capabilities are shown. There is no significant differences between \textsc{BlenderBot3} and \textsc{BlenderBot3-M\^{}3}.
    }
\end{table*}

In this section, we describe the experimental setups in detail and show results of our model \textsc{BlenderBot3-M\^{}3}, which is multi-task trained with memory management data.

\subsection{Implementation Details} 
We leverage self-reproduced \textsc{BlenderBot3} based on huggingface library as the LM backbone. 
The proposed model \textsc{BlenderBot3-M\^{}3} is multi-task trained with constructed memory management data along with data for other conversation skills, starting from the \textsf{r2c2} checkpoint. 
We use the Adam optimizer~\citep{Kingma2014AdamAM} with a cosine learning rate 5e-5 and batch size 64.
For comparison purposes, we additionally fine-tuned \textsc{BlenderBot3} on the memory management dataset. 
Experiments are performed with \textsc{BlenderBot3 3B}, \textsc{BlenderBot3 3B} + MM fine-tuning, and \textsc{BlenderBot3-M\^{}3 3B}.

\subsection{Training Dataset} 
In addition to the existing 67 training datasets of 11 particular tasks to train \textsc{BlenderBot3}, Memory Management (MM) dataset is built following the creation method introduced in Section~\ref{sec3:creation_method}. 
The created MM dataset consists of 90,000 examples, and its operation labels ($\mathrm{PASS}$, $\mathrm{APPEND}$, $\mathrm{REPLACE \ m_i}$) are equally distributed.

\subsection{Evaluation Dataset} 
The baseline model and proposed models are evaluated on Multi Session Chat (MSC) in an end-to-end manner for measuring overall conversation ability in long-term conversation, and on all 67 datasets corresponding to the training datasets for measuring performance of each task (module).

\subsection{Experimental Results}
\paragraph{End-to-End Evaluation}
The comparison of overall conversation ability on MSC dataset which has long dialogues is shown in Table~\ref{tab1:e2e-eval}. 
Models are evaluated in an end-to-end manner. 
As shown in Table~\ref{tab1:e2e-eval}, throughout all sessions, we observe an overall increase in performance of \textsc{BlenderBot3-M\^{}3} compared to \textsc{BlenderBot3}, indicating a successful integration of the memory management capability followed by an improvement in overall conversation capability.
Specifically, \textsc{BlenderBot3-M\^{}3} outperforms \textsc{BlenderBot3} with a relative 4\% performance gain in terms of F1 score (average result from all sessions).
It demonstrates that keeping memory up-to-date through memory management improves general conversation performance in long-term conversations, which further can be a cornerstone of lifelong conversations.

Additionally, the numbers of entries in memory per every 100 turns is reported in Table~\ref{tab1:e2e-eval}, showing that memory management can effectively reduce external memory usage.

\paragraph{Evaluation in other tasks}
One may consider that in exchange for the memory management ability, other existing abilities might be compromised.
To deal with this concern, we directly measure the task performance of each module, which can be also inferred from Table~\ref{tab1:e2e-eval}
Perplexity of each module is reported in Table~\ref{tab2:modular-eval}. 
Across all tasks, we observe that the average PPL score of \textsc{BlenderBot3-M\^{}3} increases 0.05 from the PPL of \textsc{BlenderBot3}. 

\vspace{2mm}
\noindent The above explorations show that the overall conversational performance has been improved by successfully incorporating the new memory management capability into conversational agents.

%% file: Sections/5_conclusion.tex
\section{Conclusion}
\label{sec:conclusion}

In this paper, we propose an effortless way to improve open-domain conversation systems by integrating memory management into them.
It is effortless in that we fully leverage existing data to construct memory management data in an automated way which can be easily scaled up.
Our proposed method does not affect BB3’s performance in other tasks, and does not require additional costs for the external memory and model parameters, but rather reduces the costs.
We show that in end-to-end conversation evaluation, our proposed model \textsc{BlenderBot3-M\^{}3}, which is multi-task trained with memory management, outperforms \textsc{BlenderBot3} with a relative 4\% performance gain in terms of F1 score. 
To deal with lifelong conversations where conversation histories and memories are accumulated endlessly, keeping memory up-to-date compactly via memory management can be a promising direction. 

\paragraph{Limitations and Future Work}
While constructing memory management dataset, we comprise memories with randomly selected user information sentences.
This may cause inconsistent memories and different distributions from those encountered in the actual conversation flows.
Therefore, a careful design of the memory could be a potential avenue for further improving model performance.

Even with management, memory will constantly increase.
Accumulated large memory is not suitable for the input of LM and occupies storage.
However, our experiments only assume long conversations where the length of maximum memory is predetermined and fixed.

%% file: Sections/7_ethic.tex
\section*{Ethical Considerations}
Since the purpose of the conversation system is to interact with human, it is important to build reliable and controllable system. Also, as the proposed system stores the information of the user, protecting privacy is also important. Lastly, we will release the dataset and code for research purpose only to prevent from unintended usage of our product.
\label{sec:ethic}